\pdfoutput=1
\documentclass[11pt]{article}
\usepackage[final]{acl}
\usepackage{times}
\usepackage{latexsym}
\usepackage[T1]{fontenc}
\usepackage[utf8]{inputenc}
\usepackage{microtype}
\usepackage{inconsolata}
\usepackage{graphicx}
\usepackage{amsmath} 
\usepackage{booktabs} 
\usepackage{subcaption}
\usepackage{tabularx}
\usepackage{diagbox}
\usepackage{pifont}
\usepackage{array} 
\usepackage{multirow}
\usepackage{xcolor}
\usepackage{microtype}
\usepackage{float}
\usepackage{makecell}
\usepackage{colortbl}
\usepackage{colortbl} 
\definecolor{promptgreen}{HTML}{13765a}
\definecolor{promptred}{HTML}{c04d5a}
\newcommand{\PreserveBackslash}[1]{\let\temp=\\#1\let\\=\temp}
\newcolumntype{C}[1]{>{\PreserveBackslash\centering}p{#1}}
\newcolumntype{R}[1]{>{\PreserveBackslash\raggedleft}p{#1}}
\newcolumntype{L}[1]{>{\PreserveBackslash\raggedright}p{#1}}

\title{Culturally-Nuanced Story Generation for Reasoning in Low-Resource Languages: The Case of Javanese and Sundanese}

\author{
    Salsabila Zahirah Pranida\thanks{Equal contribution} \quad Rifo Ahmad Genadi\footnotemark[1] \quad Fajri Koto \\
    Department of Natural Language Processing, MBZUAI \\
    \texttt{\{salsabila.pranida,rifo.genadi,fajri.koto\}@mbzuai.ac.ae}\\
}

\begin{document}
\maketitle
\begin{abstract}
Culturally grounded commonsense reasoning is underexplored in low-resource languages due to scarce data and costly native annotation. We test whether large language models (LLMs) can generate culturally nuanced narratives for such settings. Focusing on Javanese and Sundanese, we compare three data creation strategies: (1) LLM-assisted stories prompted with cultural cues, (2) machine translation from Indonesian benchmarks, and (3) native-written stories. Human evaluation finds LLM stories match natives on cultural fidelity but lag in coherence and correctness. We fine-tune models on each dataset and evaluate on a human-authored test set for classification and generation. LLM-generated data yields higher downstream performance than machine-translated and Indonesian human-authored training data. We release a high-quality benchmark of culturally grounded commonsense stories in Javanese and Sundanese to support future work.
\end{abstract}

\section{Introduction}

Reasoning, the ability to draw conclusions, make inferences, and relate concepts, is a core evaluation target in recent LLM work \citep{dubey2024llama, openai2024gpt4technicalreport, openai2024gpt4ocard, almazrouei2023falconseriesopenlanguage}. Yet widely used English benchmarks such as \texttt{StoryCloze} \citep{mostafazadeh-etal-2016-corpus, mostafazadeh-etal-2017-lsdsem}, \texttt{WinoGrande} \citep{sakaguchi2021winogrande}, and \texttt{HellaSwag} \citep{zellers-etal-2019-hellaswag} encode Western norms. Because reasoning is culturally shaped, relying on machine-translated English datasets \citep{ponti-etal-2020-xcopa, lin-etal-2022-shot} risks erasing local context.

Recent datasets for medium-resource languages (e.g., Indonesian \cite{koto-etal-2024-indoculture} and Arabic \cite{sadallah2025commonsense}) add cultural grounding but mainly target sentence-level classification. Story-level commonsense, how people interpret events across narratives, remains underexplored in low-resource languages due to limited speaker access, high annotation costs, and scarce culturally relevant materials.

We address story comprehension in two underrepresented languages, Javanese and Sundanese, spoken by roughly 80M and 32M people respectively \citep{bps2025statistik,ethnologue2025}. Beyond sheer scale, both carry rich sociolinguistic systems: Sundanese encodes politeness and hierarchy phonologically, while Javanese employs elaborate speech levels \citep{wolff1982communicative}. We adopt a \texttt{StoryCloze}-style setup \citep{mostafazadeh-etal-2016-corpus, mostafazadeh-etal-2017-lsdsem}: given a four-sentence story, models either generate a plausible fifth sentence (generation) or choose the correct continuation from two options (classification).

We compare three dataset creation strategies for culturally grounded story comprehension: (1) LLM-assisted generation with culturally informed prompts, (2) machine translation from Indonesian benchmarks, and (3) native-authored stories. Each has distinct benefits, scalability, resource reuse, and authenticity, respectively. Human evaluation shows LLM outputs are culturally plausible and comparable to native-written stories.

\begin{figure}
\centering
  \includegraphics[width=1\linewidth]{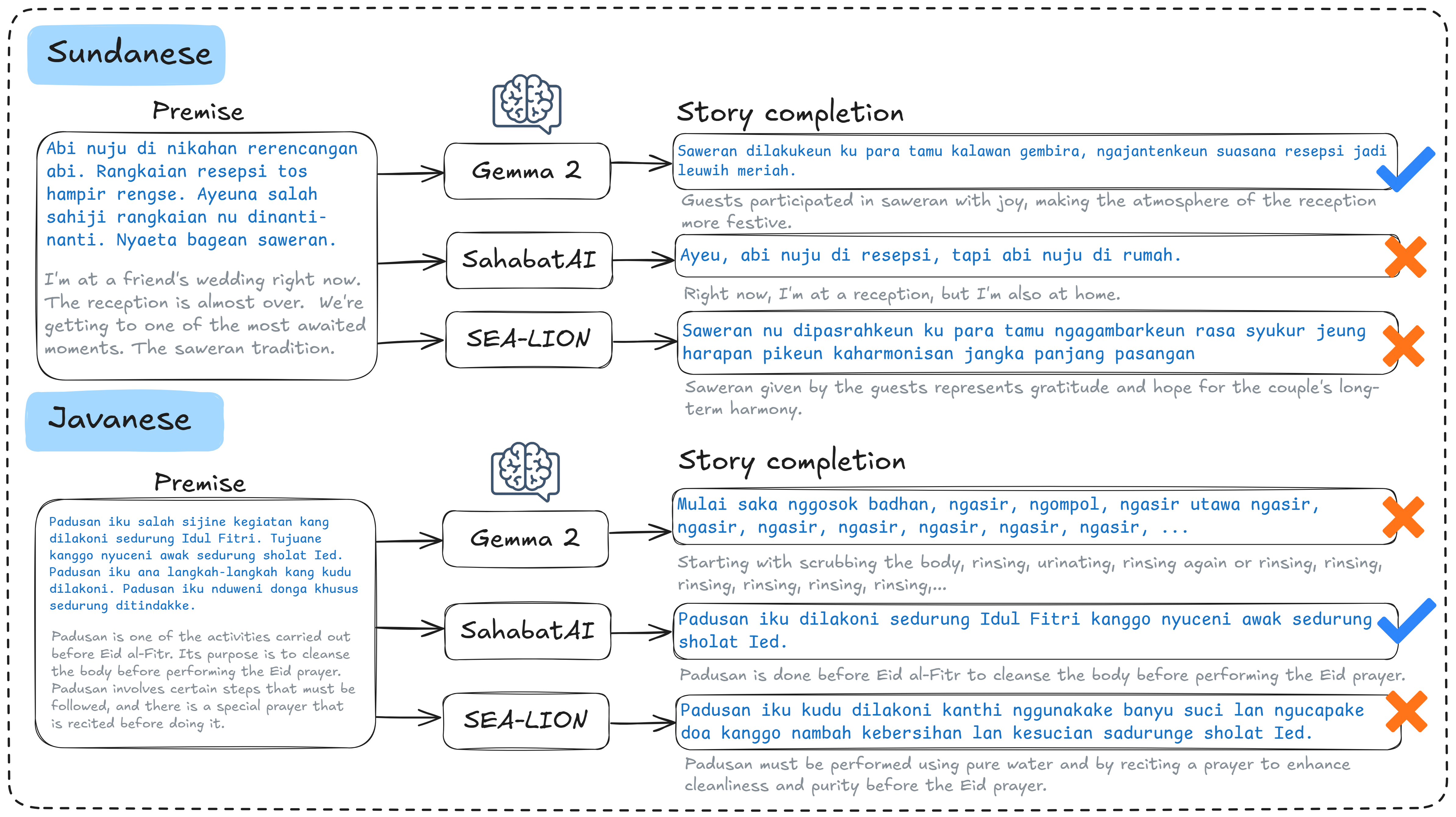}
  \caption{Examples of human-written stories in Sundanese and Javanese. English translations in gray color are provided for reference. A cross (\ding{55}) indicates a culturally irrelevant ending, generated by LLMs}
  \label{fig:teaser}
\end{figure}



To assess cultural reasoning on narratives, we use the native-authored set as a zero-shot testbed. As in Figure~\ref{fig:teaser}, Indonesian/SEA-centric models, though covering the target languages, often produce culturally inappropriate endings, showing that language support alone is insufficient. We then fine-tune on synthetic data and find that LLM-generated training yields stronger results than machine-translated data for both classification and generation.

Our contributions are as follows:
\begin{itemize}
    \item We release the \textbf{first benchmark} for culturally grounded commonsense reasoning in Javanese and Sundanese, comprising 3.3K high-quality stories. This includes 1.12K human-written samples, 1K human-reviewed machine-translated texts, and 1.22K filtered LLM-generated samples.
    \item We conduct extensive human evaluation of multiple dataset creation strategies, including (i) LLM-assisted generation, (ii) direct machine translation, (iii) culturally localized translation, and (iv) native-authored stories.
    \item We evaluate model performance through zero-shot inference and supervised fine-tuning in both classification and generation settings to assess their cultural reasoning capabilities.
\end{itemize}

\section{Related Works}
\subsection{Commonsense Reasoning in English Story Comprehension}

Story comprehension in NLP involves reasoning over causal, temporal, and commonsense relations within narratives. The \texttt{StoryCloze} test, introduced by \citet{mostafazadeh-etal-2016-corpus, mostafazadeh-etal-2017-lsdsem} is a landmark benchmark, requiring models to select the most plausible ending for a short four-sentence story. Many commonsense reasoning datasets, however, focus on sentence-level challenges include \texttt{WinoGrande} \citep{sakaguchi2021winogrande} for pronoun resolution, COPA \citep{gordon-etal-2012-semeval} for causal reasoning, and \texttt{HellaSwag} \citep{zellers-etal-2019-hellaswag} for adversarial sentence completion. While effective for probing localized reasoning, these do not capture broader discourse coherence or character motivations.

Recent work has shifted toward narrative-level reasoning with longer contexts and richer event dynamics. NarrativeQA \citep{kocisky-etal-2018-narrativeqa} covers full books and movie scripts, CosmosQA \citep{huang-etal-2019-cosmos} infers implicit causes and intentions, and TellMeWhy \citep{lal-etal-2021-tellmewhy} targets causal and motivational “why” questions. Yet these remain English-centric and question-answering-oriented. Our work instead addresses narrative completion in low-resource languages, particularly in Javanese and Sundanese, providing a culturally grounded alternative to high-resource, English-dominant benchmarks.

\subsection{Commonsense Reasoning in Languages Beyond English} 

Early multilingual commonsense benchmarks often extended English datasets via translation. \texttt{XCOPA} \citep{ponti-etal-2020-xcopa} translated COPA into 11 typologically diverse languages, including Indonesian, while \texttt{X-CSQA} \citep{lin-etal-2021-common} adapted \texttt{CommonSenseQA} across languages. Although useful for cross-lingual evaluation, such resources inherit Anglocentric biases, as progress in English does not always transfer culturally or linguistically \citep{lin-etal-2022-shot, shwartz-etal-2020-unsupervised}. Direct translations risk embedding English social contexts rather than local commonsense \citep{lin-etal-2021-common}.

Story comprehension tasks like \texttt{StoryCloze} \citep{mostafazadeh-etal-2016-corpus} have been similarly extended. One such extenstion is \texttt{XStoryCloze} \citep{lin-etal-2022-shot}, by translating English narratives into multiple languages. Yet such approaches still struggle to capture culture-specific narrative norms.

For Indonesian, culturally grounded datasets such as \texttt{COPAL-ID} \citep{wibowo-etal-2024-copal} and \texttt{IndoCulture} \citep{koto-etal-2024-indoculture} model regional practices and norms across 11 provinces, advancing evaluation in a medium-resource language. However, they primarily focus on short-form, sentence-level reasoning such as multiple-choice or cloze-style questions, rather than full-narrative comprehension. Beyond Indonesia, \texttt{CultureBank} \citep{shi-etal-2024-culturebank} compiles large-scale cultural knowledge from community narratives to support culturally aware language technologies, while \texttt{CultureLLM} \citep{li-etal-2024-culturellm} incorporates cultural differences into LLMs via semantic data augmentation. However, these resources primarily focus on short-form, structured tasks rather than full-narrative comprehension. Our work fills this gap by introducing the first benchmark for \textbf{story-level commonsense reasoning} in low-resource languages, specifically Javanese and Sundanese, two of Indonesia’s most widely spoken local languages.


\subsection{LLM-Generated Data Creation}

One possible solution to tackle data scarcity in
NLP is applying data augmentation \citep{feng-etal-2021-survey, ding-etal-2020-daga, ahmed-buys-2024-neural, liu2024best, yong-etal-2024-lexc, guo2024generativeaisyntheticdata, liu-etal-2022-wanli}, with LLMs increasingly used to produce high-quality synthetic data that complements or substitutes manual annotation. Most prior work targets classification tasks. For example, WANLI \citep{liu-etal-2022-wanli} used GPT-3 \citep{brown2020language} to generate synthetic English natural language inference data \citep{bowman-etal-2015-large} refined by humans, while \citet{yong-etal-2024-lexc} generated English sentiment and topic classification data before translating it into low-resource languages using bilingual lexicons.

For low-resource languages, \citet{putri-etal-2024-llm} employed GPT-4 \citep{openai2024gpt4technicalreport} to create question-answering datasets, showing LLM potential in under-resourced settings. However, such efforts often overlook cultural reasoning and narrative coherence. Our work instead focuses on culturally nuanced story generation in Javanese and Sundanese, targeting story-level commonsense reasoning. We compare multiple data creation strategies, including LLM-assisted generation with open- and closed-weight models, machine translation from Indonesian, and native-authored stories.

\section{Dataset Construction}
\begin{figure}[ht] 
    \centering
    \includegraphics[width=0.5\textwidth]{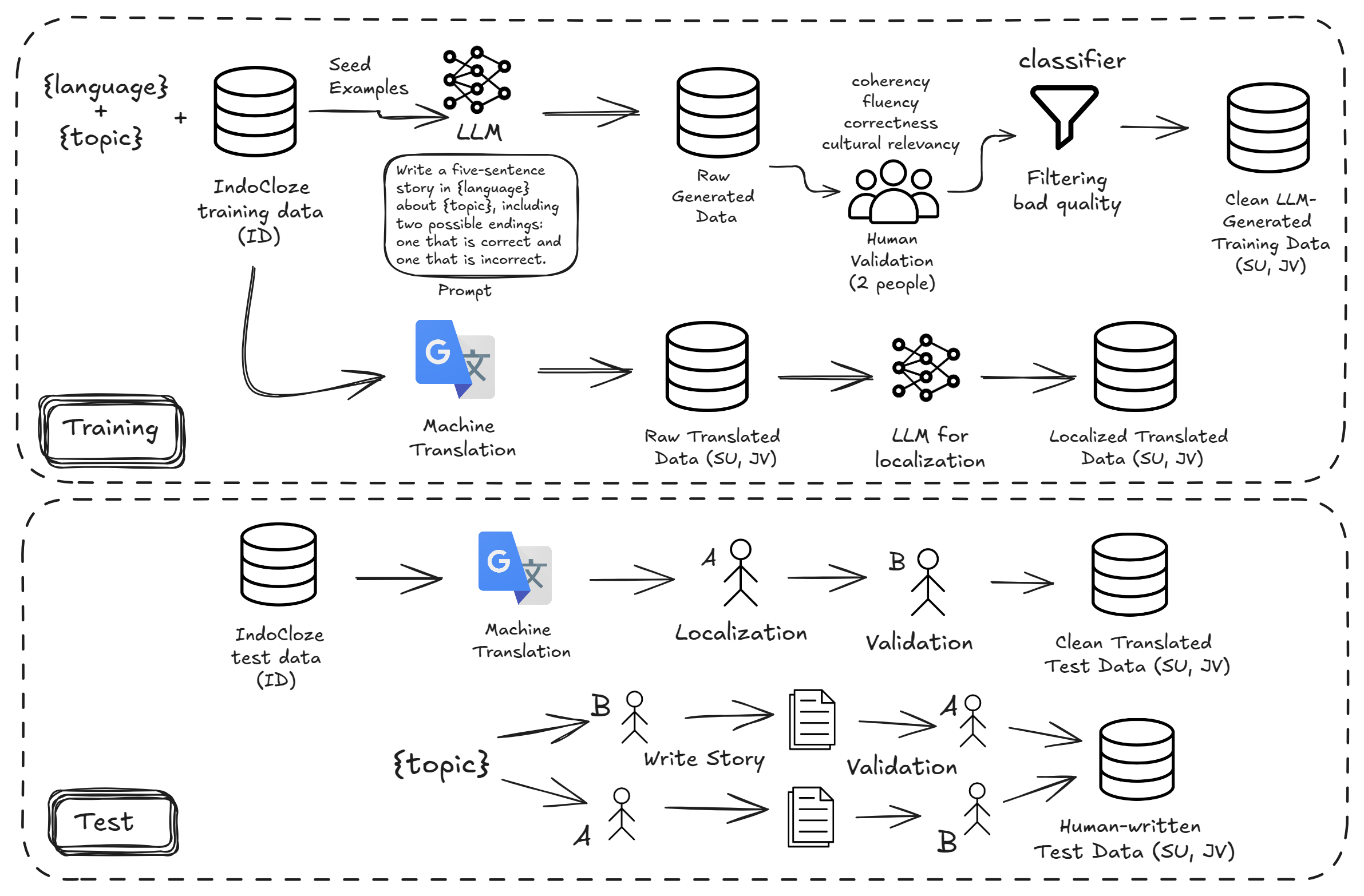} 
    \caption{Overall pipeline of dataset creation.}
    \label{fig:local_cloze_pipeline}
\end{figure}

\label{sect:dataset_construction}



As in Figure~\ref{fig:local_cloze_pipeline}, we build two parallel streams: training and test. For training, the \texttt{IndoCloze} \citep{koto-etal-2022-cloze} train split serves both as seeds for LLM-guided generation and as sources for machine translation into Javanese and Sundanese. For testing, we translate the \texttt{IndoCloze} test split and human-verify it for linguistic quality and cultural relevance, and we add a fully new set of native-authored stories from predefined topics. Each instance follows the StoryCloze format: a four-sentence premise with one correct and one incorrect ending.

Indonesian is chosen as the seed language for its national status and cultural proximity to Javanese and Sundanese. To ensure authenticity, native speakers authored and validated stories, embedding local names, places, foods, and customs. Following \citet{mostafazadeh-etal-2016-corpus,koto-etal-2022-cloze}, the dataset targets everyday commonsense reasoning grounded in local culture

\subsection{Training Data}  

We construct our training set using three strategies: (1) LLM-assisted data generation, (2) direct machine translation, and (3) machine translation followed by cultural localization using an LLM.


\subsubsection{LLM-Assisted Data Generation}
We synthesize training data with three open models: \texttt{Gemma2-27B-it} \citep{team2024gemma}, \texttt{Llama3.1-70B} \citep{dubey2024llama}, \texttt{Mixtral-8x7B-Instruct} \citep{jiang2024mixtral}, and three closed models: \texttt{GPT-4o} \citep{openai2024gpt4ocard, openai2024gpt4technicalreport}, \texttt{Cohere Command-R-Plus} \citep{cohere2024commandrplus}, \texttt{Claude-3-Opus} \citep{anthropic2024claude3}.


For each language and LLM, we supply seed examples and topics. We manually translate 50 \texttt{IndoCloze} train samples into Javanese and Sundanese with cultural localization (e.g., foods like \textit{Gudeg} and rituals like \textit{Sawéran}). Topics are derived from \texttt{IndoCulture} \citep{koto-etal-2024-indoculture}. The full prompt appears in Figure~\ref{fig:prompt_template}.

To ensure that the generated examples align with the intended cultural context, we use cultural topics derived from \texttt{IndoCulture} \citep{koto-etal-2024-indoculture}. The full prompt used in this study is provided in Figure~\ref{fig:prompt_template}.



Each LLM produces 2,000 stories (1,000 per language), yielding six training sets. Per call, five random seeds guide generation at temperature 0.7; we repeat until each model reaches 1,000 valid samples per language ($\approx166$ per topic on average).

To filter quality, we train an XLM-R binary classifier on 600 human-rated outputs (Section~\ref{sec:quality_analysis})\footnote{68.33\% accuracy on the dev split (20\% of train); setup in Appendix~\ref{sec:training_configurations}.}. Applied to all 12,000 generations, it retains 1.2K high-quality stories (592 Sundanese, 628 Javanese). Roughly half of Claude and GPT-4o outputs pass, while far fewer from other LLMs, especially Mixtral, survive (Appendix~\ref{sec:filtered_data_proportion}).


\subsubsection{Machine Translation (\textbf{MT\textsubscript{train}})}
We translate 1,000 \texttt{IndoCulture} training instances into Javanese and Sundanese (1,000 each) using Google Translate\footnote{\url{https://translate.google.com/}, accessed September 2024.} (see Appendix~\ref{sec:machine_translation_quality_comparison} for quality). As a training resource, this set is \emph{not} human-validated. We refer to it as \textbf{MT\textsubscript{train}}.


\subsubsection{Machine Translation + Localization  (\textbf{MT\textsubscript{train}+GPT4o)}} 
We prompt GPT-4o to culturally localize the MT outputs, following the same guidelines given to human annotators. The model adapts names, events, foods, settings, and social norms to Javanese/Sundanese contexts. This probes whether LLMs can align literal translations with local cultural values to produce more authentic, context-appropriate narratives.

\subsection{Test Set}
\label{subsect:llm_generated_data_construction}

We carefully construct the test set using two strategies: (1) machine translation with human verification, resulting in 500 Javanese and 500 Sundanese instances, and (2) manual writing by native speakers based on pre-defined topics, producing roughly 529 Javanese instances and 595 Sundanese instanecs. Each strategy undergoes rigorous quality control to ensure accuracy and reliability. In total, we create a high-quality test set of 2,124 instances across both languages.

To ensure the authenticity and quality of the dataset, we recruited 4 expert workers (2 per language) who are not only native Indonesian speakers but also fluent in Javanese and Sundanese. Each expert worker has a deep understanding of their respective language, culture, and customs. They have at least 10 years of experience speaking Javanese or Sundanese and possess strong linguistic and cultural expertise. The recruited workers, aged between 21 and 35 years, hold bachelor's degrees and were carefully selected for their proficiency in both language and cultural knowledge.

\subsubsection{Machine translation with human verification (\textbf{MT\textsubscript{test}+Human})} As shown in Figure~\ref{fig:local_cloze_pipeline}, we translate 500 randomly selected samples from the \texttt{IndoCloze} test set into Javanese and Sundanese using Google Translate.\footnote{Google Translate was accessed in September 2024} To ensure accuracy and naturalness of the machine-translation, we employ two native speakers for each language and implement a two-stage quality control process. In Stage 1, the first worker manually corrects translation errors and localizes content by replacing entities (e.g., names, buildings, food) with culturally relevant alternatives.\footnote{Note that while we applied cultural localization, not all examples could be fully adapted to Javanese or Sundanese contexts, as some stories reflect general Indonesian cultural elements that are not specific to either group.} In Stage 2, a second worker validates the revised text and directly corrects any remaining errors from the first stage. From this point forward, we refer to this data as \textbf{MT\textsubscript{test}+Human}.

\subsubsection{Human-written Dataset (\textbf{HW})} Each expert worker in Sundanese and Javanese is tasked with writing 600 short stories following the \texttt{IndoCloze} format: a four-sentence premise, a correct fifth sentence, and an incorrect fifth sentence. Stories are written based on 12 predefined topics, adhering to the same topic taxonomy used for training (see Section~\ref{subsect:llm_generated_data_construction}).See Appendix~\ref{sec:story_guideline} for further details on the writing guidelines.

To ensure quality, each expert worker reviewed their peer’s written stories. The reviewing worker was presented with a premise and two randomized alternate endings from another worker’s story and was asked to identify the correct one. Instances incorrectly identified by the second worker were discarded, as they likely contained incorrect endings or exhibited ambiguity. After quality control, 529 Javanese and 595 Sundanese instances remained from the original 600 per language. From this point forward, we refer to this human-written dataset as \textbf{HW}.

\section{Data Analysis}
\subsection{Overall Statistics}
For LLM-assisted dataset creation, GPT-4o and Claude demonstrated the highest efficiency, generating nearly 1,000 clean samples with minimal discarded output, while Mixtral was the least efficient, requiring significantly more samples to reach the same threshold. The LLM-generated data does not have a uniform topic distribution due to variations in broken samples.

In total, the LLM-generated datasets contain 72K sentences and approximately 557K words. The word distribution across sentence positions remains consistent across the six LLMs, with word counts per position relatively uniform and a median sentence length ranging from 5 to 10 words. MT\textsubscript{test}+Human (both with and without localization) and HW datasets exhibit a similar word distribution pattern to the LLM-generated datasets. MT contains around 6K sentences with 44,5K words, while HW has 6,7K sentences with 58,2K words. Despite the slight difference in word count, both datasets maintain a consistent distribution, with a median sentence length ranging from 4 to 11 words.

\begin{table*}[ht!]
\small
\centering
\caption{Quality analysis of models on Sundanese and Javanese. Higher scores indicate better performance in each category.}
\resizebox{0.8\linewidth}{!}{%
\begin{tabular}{lcccccccc}
\toprule
\multirow{2}{*}{\textbf{Dataset}} & \multicolumn{4}{c}{\cellcolor{blue!7}\textbf{Sundanese}} & \multicolumn{4}{c}{\cellcolor{red!7}\textbf{Javanese}} \\
 & \cellcolor{blue!7}\textbf{Coherence} & \cellcolor{blue!7}\textbf{Fluency} & \cellcolor{blue!7}\textbf{Correctness} & \cellcolor{blue!7}\textbf{Cultural Rel.} & \cellcolor{red!7}\textbf{Coherence} & \cellcolor{red!7}\textbf{Fluency} & \cellcolor{red!7}\textbf{Correctness} & \cellcolor{red!7}\textbf{Cultural Rel.} \\ \midrule
Human Written (HW) & \textbf{5.0} & \textbf{5.0} & \textbf{100} & \textbf{96} & \textbf{5.0} & \textbf{5.0} & \textbf{100} & 66 \\
\midrule
\multicolumn{9}{c}{\cellcolor{gray!15}\textbf{LLM Generated Data}} \\
GPT-4o & 4.7 & 4.2 & 80 & \textbf{96} & 4.9 & 4.5 & 97 & 91 \\
Claude & 4.7 & 4.4 & 86 & 92 & 4.9 & 4.3 & 96 & \textbf{93} \\
Cohere & 3.4 & 3.0 & 28 & 46 & 4.6 & 4.1 & 80 & 65 \\
Llama3.1 & 3.7 & 3.4 & 56 & 70 & 4.5 & 4.2 & 65 & 50 \\
Gemma2 & 3.9 & 3.1 & 42 & 78 & 4.8 & 3.6 & 83 & 81 \\
Mixtral & 2.0 & 1.5 & 0 & 4 & 2.0 & 1.9 & 3 & 22 \\ \midrule
\multicolumn{9}{c}{\cellcolor{gray!15} \textbf{Translated}} \\
MT\textsubscript{train} & 3.24	&4.36&	80&	20&	4.36	&4.46&	98&	12 \\
\midrule
\multicolumn{9}{c}{\cellcolor{gray!15} \textbf{Translated Data + Localization}} \\
MT\textsubscript{train} + GPT4o  & 4.36&	4.68&	86	&80&	4.6&	4.64&	98&	76 \\
MT\textsubscript{test} + Human & \textbf{5.0} & \textbf{5.0} & \textbf{100} & 14 & \textbf{5.0} & \textbf{5.0} & \textbf{100} & 18 \\

\bottomrule
\end{tabular}%
}
\label{tab:quality_analysis}
\end{table*}

\begin{table*}[ht!]
\small
\centering
\caption{Lexical diversity analysis of different models on Sundanese and Javanese. ``LW'' denotes the percentage of loanword.}
\resizebox{0.7\linewidth}{!}{%
\begin{tabular}{lcccccccc}
\toprule
\multirow{2}{*}{\textbf{Dataset}} & \multicolumn{4}{c}{\cellcolor{blue!7}\textbf{Sundanese}} & \multicolumn{4}{c}{\cellcolor{red!7}\textbf{Javanese}} \\
 & \cellcolor{blue!7}\textbf{\#data} & \cellcolor{blue!7}\textbf{\#vocab} & \cellcolor{blue!7}\textbf{LW (\%) $\downarrow$} & \cellcolor{blue!7}{\textbf{MATTR} $\uparrow$} & \cellcolor{red!7}\textbf{\#data} & \cellcolor{red!7}\textbf{\#vocab} & \cellcolor{red!7}\textbf{LW (\%) $\downarrow$} & \cellcolor{red!7}{\textbf{MATTR} $\uparrow$} \\ \midrule
 Human Written (HW) & 594 & 4693 & \textbf{0} & \textbf{0.84} & 529 & 3497 & \textbf{0} & \textbf{0.80} \\
\midrule
\multicolumn{9}{c}{\cellcolor{gray!15} \textbf{LLM Generated Data}} \\
GPT-4o & 1000 & 3444 & \textbf{0} & 0.82 & 1000 & 3073 & \textbf{0} & \textbf{0.80} \\
Claude & 1000 & 3898 & \textbf{0} & 0.80 & 1000 & 3104 & \textbf{0} & 0.79 \\
Cohere & 1000 & 2654 & 3 & 0.65 & 1000 & 2254 & 3 & 0.68 \\
Llama3.1 & 1000 & 3126 & 0 & 0.69 & 1000 & 2836 & 2 & 0.69 \\
Gemma2 & 1000 & 3758 & 5 & 0.72 & 1000 & 3334 & 4 & 0.71 \\
Mixtral & 1000 & 4584 & {16} & 0.71 & 1000 & 4215 & {13} & 0.67 \\ 
\midrule
\multicolumn{9}{c}{\cellcolor{gray!15} \textbf{Translated Data}} \\
MT\textsubscript{train} & 1000 &	5272&	\textbf{0} &	0.83&	1000&	5007&	1&	\textbf{0.81} \\

\midrule
\multicolumn{9}{c}{\textbf{\cellcolor{gray!15} Translated Data + Localization}} \\
MT\textsubscript{train} + GPT4o & 1000&	4985&	1&	0.82&	1000&	4692&	1	& \textbf{0.81}  \\ 
MT\textsubscript{test} + Human & 500 & 3877 & \textbf{0} & 0.82 & 500 & 3620 & 3 & \textbf{0.81} \\
\bottomrule
\end{tabular}%
}
\label{tab:lexical_diversity}
\end{table*}

\subsection{Quality Analysis based on Human Evaluation}
\label{sec:quality_analysis}
We evaluate the quality of all the constructed dataset based on four key criteria: coherence, fluency, correctness, and cultural relevance. Specifically, we randomly select 50 samples (5–10\% of the total dataset) for both Sundanese and Javanese and engage two native speakers for evaluation. Coherence and fluency are rated on a Likert scale from 0 to 5, while correctness and cultural relevance are assessed using binary annotation and reported as percentages. More details on the annotation guideline can be found in Appendix~\ref{sec:worker_guideline}. Table~\ref{tab:quality_analysis} summarizes the results of human evaluation, with scores for coherence and fluency being averaged between the annotators. Meanwhile, for correctness and cultural relevance, we count the percentage of data perceived as correct or culturally relevant by both annotators. Inter-annotator agreement scores ranges from 0.4 to 0.7, as shown in Appendix~\ref{sec:agreement_score}.

We observe that among the evaluated LLMs, both GPT-4o and Claude consistently demonstrate strong performance across all metrics. Notably, their cultural relevance scores are comparable to those of human-written texts. Applying localization to the machine-translated data using GPT-4o (MT$_\text{train}$+GPT-4o) improves coherence and cultural relevance. Finally, while human post-editing ensures near-perfect scores in coherence, fluency, and correctness, achieving full cultural localization remains challenging, as not all content can be naturally adapted without compromising narrative plausibility.


\subsection{Lexical Diversity}
We analyze the lexical diversity of LLM-generated data for Sundanese and Javanese using key metrics such as vocabulary size, moving-average type-token ratio (MATTR) \cite{covington2010cutting}, and the proportion of loanwords. To measure loanword presence, we manually review the top 100 most frequent words for each model.

As shown in Table~\ref{tab:lexical_diversity}, GPT-4o achieves a high MATTR score, making it highly comparable to human-written data. Among the LLM-generated datasets, Mixtral has the largest vocabulary size but also introduced the highest proportion of loanwords, with 16\% in Sundanese and 13\% in Javanese, suggesting a significant reliance on non-native terms. In contrast, GPT-4o and Claude generate text entirely in Sundanese and Javanese without incorporating foreign words, highlighting their ability to produce better datasets. 

Upon manual inspection, we find that LLMs frequently generate common Javanese names such as \textit{Ayu}, \textit{Dwi}, \textit{Bayu}, \textit{Eko}, and \textit{Sari}. For Sundanese, commonly produced names include \textit{Lia}, \textit{Budi}, \textit{Dewi}, and \textit{Rina}. While these names are widely used across Indonesia, the variation in honorific terms is limited across all models, only \textit{Pak} and \textit{Bu}, along with their formal variants \textit{Bapak} and \textit{Ibu}, appear consistently. This suggests that, although some models demonstrate surface-level lexical diversity, deeper sociolinguistic features such as honorific variation remain underrepresented.

\section{Experiments and Analysis}
\subsection{Classification}
\subsubsection{Setup} 
We adopt the classification accuracy metric, as proposed in both \citet{mostafazadeh-etal-2016-corpus} and \citet{koto-etal-2022-cloze}, defined as the ratio of correctly predicted instances to the total number of test cases.
For our experiments, we fine-tune several models: \texttt{Qwen 2.5 7B}~\citep{qwen2025qwen25technicalreport}, \texttt{Llama 3.1 8B}~\citep{grattafiori2024llama3herdmodels}, \texttt{Gemma 2 9B}~\citep{team2024gemma}, \texttt{SahabatAI Llama 8B}~\citep{goto2024sahabatai}, \texttt{SEA-LION Llama 8B}~\citep{ng2025sealionsoutheastasianlanguages}, and \texttt{XLM-R}~\citep{abs-1911-02116}. For LLMs, we conduct instruction fine-tuning with a multiple-choice question framework using a LoRA adapter \cite{hu2022lora} on the combined Javanese and Sundanese training set. Full training details are provided in Appendix~\ref{sec:training_configurations}. To ensure robustness, results are averaged over three runs. For comparison, we also report the zero-shot performance of each model prior to fine-tuning.

In addition to the zero-shot setting, we experiment with the following training data variations: (i) MT$_\text{train}$ (machine-translated data), (ii) MT$_\text{train}$+GPT-4o (machine-translated data localized using GPT-4o), (iii) All LLM (the full set of LLM-generated samples), and (iv) LLM\_Filt (a filtered subset comprising the top 10\% of All LLM based on human evaluation). We evaluate each model on two test sets: (i) a human-written set and (ii) a machine-translated set.

\begin{table*}[ht]
\centering
\caption{Accuracy of models on the human-written test set in Javanese and Sundanese across training sets.}
\label{tab:human_written_accuracy}
\resizebox{0.8\linewidth}{!}{%
\small
\begin{tabular}{lcccccccccc}
\toprule
\rowcolor{white}
\textbf{Model} & \multicolumn{5}{c}{\cellcolor{green!15}\textbf{Javanese}} & \multicolumn{5}{c}{\cellcolor{yellow!20}\textbf{Sundanese}} \\
\rowcolor{white}
& \cellcolor{green!15}\textbf{0-shot} 
& \cellcolor{green!15}\textbf{MT$_\text{Train}$} 
& \cellcolor{green!15}\textbf{MT$_\text{Train}$+GPT-4o} 
& \cellcolor{green!15}\textbf{All\_LLM} 
& \cellcolor{green!15}\textbf{LLM\_Filt} 
& \cellcolor{yellow!20}\textbf{0-shot} 
& \cellcolor{yellow!20}\textbf{MT$_\text{Train}$} 
& \cellcolor{yellow!20}\textbf{MT$_\text{Train}$+GPT-4o} 
& \cellcolor{yellow!20}\textbf{All\_LLM} 
& \cellcolor{yellow!20}\textbf{LLM\_Filt} \\
\midrule
XLM-R & {NA} & 69.00 & 67.67 & 80.72 & \textbf{81.47} & {NA} & 75.42 & 50.34 & \textbf{75.93} & 72.73 \\
Qwen 2.5 & 76.94 & 85.44 & 84.31 & 84.88 & \textbf{87.71} & 64.48 & 74.75 & 72.22 & \textbf{80.64} & 79.63 \\
Llama 3.1 & 75.24 & 90.17 & 86.96 & 91.68 & \textbf{92.44} & 48.32 & 79.97 & 71.55 & \textbf{81.14} & 78.62 \\
Gemma 2 & 85.07 & 93.95 & \textbf{95.46} & \textbf{95.46} & 95.27 & 51.52 & 86.20 & \textbf{89.23} & 87.88 & 85.02 \\
SahabatAI Llama & 77.13 & 95.46 & 94.14 & 97.16 & \textbf{97.73} & 34.51 & 87.88 & 87.21 & 91.25 & \textbf{92.42} \\
SEA-LION Llama & 69.19 & 96.60 & 94.71 & 93.76 & \textbf{95.27} & 25.93 & 88.55 & 83.67 & \textbf{82.15} & 77.44 \\
\bottomrule
\end{tabular}%
}
\end{table*}

\begin{table*}[ht]
\centering
\caption{Accuracy of models on the machine-translated test set in Javanese and Sundanese across training sets.}
\label{tab:mt_test_accuracy}
\resizebox{0.8\linewidth}{!}{%
\small
\begin{tabular}{lcccccccccc}
\toprule
\rowcolor{white}
\textbf{Model} & \multicolumn{5}{c}{\cellcolor{green!15}\textbf{Javanese}} & \multicolumn{5}{c}{\cellcolor{yellow!20}\textbf{Sundanese}} \\
\rowcolor{white}
& \cellcolor{green!15}\textbf{0-shot} 
& \cellcolor{green!15}\textbf{MT$_\text{Train}$} 
& \cellcolor{green!15}\textbf{MT$_\text{Train}$+GPT-4o} 
& \cellcolor{green!15}\textbf{All\_LLM} 
& \cellcolor{green!15}\textbf{LLM\_Filt} 
& \cellcolor{yellow!20}\textbf{0-shot} 
& \cellcolor{yellow!20}\textbf{MT$_\text{Train}$} 
& \cellcolor{yellow!20}\textbf{MT$_\text{Train}$+GPT-4o} 
& \cellcolor{yellow!20}\textbf{All\_LLM} 
& \cellcolor{yellow!20}\textbf{LLM\_Filt} \\
\midrule
XLM-R & {NA} & 69.40 & \textbf{70.20} & 66.80 & 62.60 & {NA} & 56.60 & \textbf{73.00} & 67.60 & 66.00 \\
Qwen 2.5 & 68.00 & \textbf{80.00} & 75.00 & 75.80 & 75.80 & 63.60 & 78.80 & 74.20 & 76.60 & \textbf{77.60} \\
Llama 3.1 & 47.60 & \textbf{86.00} & 70.40 & 77.40 & 73.40 & 48.60 & \textbf{83.40} & 74.00 & 75.00 & 76.00 \\
Gemma 2 & 63.00 & 91.80 & \textbf{92.80} & 83.60 & 83.40 & 57.60 & 90.00 & \textbf{91.80} & 80.80 & 83.40 \\
SahabatAI Llama & 34.80 & 90.00 & 87.60 & 84.20 & \textbf{88.40} & 30.80 & 87.40 & 86.60 & 83.60 & \textbf{87.60} \\
SEA-LION Llama & 32.20 & \textbf{92.60} & 87.40 & 76.20 & 72.20 & 29.80 & \textbf{92.20} & 88.80 & 77.60 & 71.20 \\
\bottomrule
\end{tabular}%
}
\end{table*}

\subsubsection{Overall Performance} 
Table~\ref{tab:human_written_accuracy} and Table~\ref{tab:mt_test_accuracy} present the classification accuracies of all models trained on different training sets, evaluated on the human-written and machine-translated test sets, respectively. In the zero-shot setting, model performance ranges from approximately 34\% to 70\% across both Javanese and Sundanese, underscoring the inherent challenges of cultural commonsense reasoning in these languages. Fine-tuning consistently improves model performance over the zero-shot baseline, with LLMs showing particularly strong gains. Notably, XLM-R underperforms relative to the LLMs, suggesting that large generative models are more effective at capturing cultural nuances. 

We observe that models trained on LLM-generated data perform best when evaluated on the human-written test set (Table~\ref{tab:human_written_accuracy}), while those trained on machine-translated data tend to perform better on the machine-translated test set (Table~\ref{tab:mt_test_accuracy}), indicating some sensitivity to data distribution alignment. Interestingly, localizing the machine-translated training data (MT$_\text{train}$+GPT-4o) does not consistently lead to improved model performance compared to using the original machine-translated data (MT$_\text{train}$). 

\begin{figure}[h]
    \centering
    \includegraphics[width=.7\linewidth]{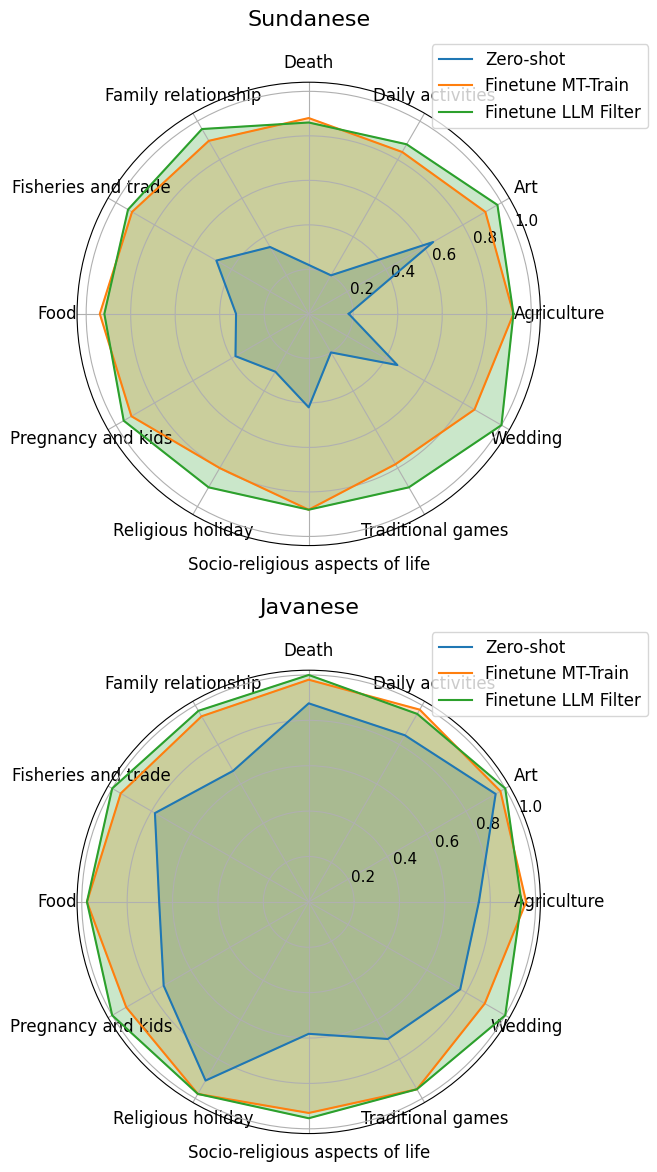}
\caption{Topic-wise accuracy of SahabatAI (an Indonesian-centric LLM) on the human-written test set for Javanese and Sundanese, comparing zero-shot, and fine-tuning on MT$_\text{train}$fine-tuned, and LLM$_\text{filtered}$.}
    \label{fig:radarchart-topic}
    \vspace{-0.5cm}
\end{figure}

\subsubsection{Performance Across Different Topics} 
Figure~\ref{fig:radarchart-topic} presents the topic-wise performance of \texttt{SahabatAI} on the human-written test set. The chart shows that models fine-tuned on both MT\textsubscript{train} and LLM$_\text{Filtered}$ consistently achieve higher accuracy across most categories compared to the zero-shot baseline. Interestingly, the zero-shot setting performs relatively well on the \textit{Art} category. Fine-tuning on LLM$_\text{Filtered}$ yields notable improvements in culturally rich topics such as \textit{Wedding}, \textit{Pregnancy}, and \textit{Art}, outperforming models trained on MT\textsubscript{train}.


\subsection{Generation}
\subsubsection{Setup} Using the same set of LLMs from the classification experiments, we fine-tune the models to perform story continuation: given a four-sentence premise, the model is trained to generate a coherent fifth sentence in either Javanese or Sundanese. We apply supervised fine-tuning using QLoRA \citep{dettmers2023qlora}, with training details provided in Appendix~\ref{sec:training_configurations}. To evaluate generation quality, we use two automatic metrics: ROUGE-L \citep{lin-2004-rouge} for lexical overlap and BERTScore \citep{zhang2019bertscore} for semantic similarity.

\subsubsection{Overall Performance} Table~\ref{tab:human_written_bertscore_rl} and Table~\ref{tab:mt_test_bertscore_rl} presents the automated evaluation metrics for different training sets across Sundanese and Javanese on both human-written and machine-translated test sets. Finetuning consistently improves performance over the 0-shot. Notably, on the human-written data that contains more cultural nuanced story, model fine-tuned with LLM-generated data gives higher improvement compared to others.

Among the models, SEA-LION Llama achieves the highest score across most test, then followed closely by Gemma 2. The LLM\_Filt data often matches or outperforms the All\_LLM settings that uses more samples. Similar to classification, in human-written test set, LLM-generated-training data tends to be better than in machine-translated test set.

\begin{table*}[t]
\centering
\caption{BERTScore / ROUGE-L F1 of models on the human-written test set in Javanese and Sundanese across training sets.}
\label{tab:human_written_bertscore_rl}
\resizebox{0.85\linewidth}{!}{%
\small
\begin{tabular}{lcccccccccc}
\toprule
\rowcolor{white}
\textbf{Model} & \multicolumn{5}{c}{\cellcolor{green!15}\textbf{Javanese}} & \multicolumn{5}{c}{\cellcolor{yellow!20}\textbf{Sundanese}} \\
\rowcolor{white}
& \cellcolor{green!15}\textbf{0-shot} 
& \cellcolor{green!15}\textbf{MT} 
& \cellcolor{green!15}\textbf{MT$_\text{Train}$+GPT-4o}  
& \cellcolor{green!15}\textbf{All\_LLM} 
& \cellcolor{green!15}\textbf{LLM\_Filt} 
& \cellcolor{yellow!20}\textbf{0-shot} 
& \cellcolor{yellow!20}\textbf{MT$_\text{Train}$} 
& \cellcolor{yellow!20}\textbf{MT$_\text{Train}$+GPT-4o} 
& \cellcolor{yellow!20}\textbf{All\_LLM} 
& \cellcolor{yellow!20}\textbf{LLM\_Filt} \\
\midrule
Qwen 2.5 & 72.5 / 21.8 & 71.7 / 19.2 & 72.0 / 19.5 & \textbf{72.4 / 21.7} & 71.9 / 19.5 
     & 69.2 / 15.4 & 70.3 / 15.6 & \textbf{70.3 / 16.5} & \textbf{70.5 / 16.7} & 69.2 / 14.6 \\
Llama 3.1 & 62.4 / 8.3 & 72.7 / 21.5 & 72.4 / 21.1 & \textbf{72.7 / 22.9} & 72.7 / 22.2 
      & 59.4 / 5.6 & 70.0 / 16.1 & 70.3 / 16.23 & \textbf{70.6 / 17.9} & 68.9 / 15.5 \\
Gemma 2 & 70.9 / 17.7 & \textbf{73.2 / 24.0} & 73.1 / 23.1 & 72.8 / 22.9 & 72.6 / 22.2 
      & 68.2 / 13.2 & \textbf{70.3 / 18.1} & 70.2 / 17.9 & 70.8 / 17.9 & 70.2 / 17.1 \\
SahabatAI Llama & 63.1 / 15.3 & \textbf{71.8 / 20.1} & 59.7 / 9.2 & 62.6 / 13.6 & 67.2 / 16.6 
               & 57.7 / 8.9 & 69.2 / 16.2 & 59.1 / 7.5 & 62.4 / 11.7 & \textbf{70.6 / 17.6} \\
SEA-LION Llama & \textbf{72.7 / 23.0} & 72.4 / 20.9 & 72.5 / 21.5 & \textbf{72.8 / 23.9} & 72.6 / 22.5 
               & 68.9 / 16.7 & 70.1 / 16.3 & 70.3 / 17.0 & 71.0 / 18.3 & \textbf{71.3 / 18.6} \\

\bottomrule
\end{tabular}%
}
\end{table*}

\begin{table*}[t]
\centering
\caption{BERTScore / ROUGE-L F1 of models on the machine-translated test set in Javanese and Sundanese across training sets.}
\label{tab:mt_test_bertscore_rl}
\resizebox{0.85\linewidth}{!}{%
\small
\begin{tabular}{lcccccccccc}
\toprule
\rowcolor{white}
\textbf{Model} & \multicolumn{5}{c}{\cellcolor{green!15}\textbf{Javanese}} & \multicolumn{5}{c}{\cellcolor{yellow!20}\textbf{Sundanese}} \\
\rowcolor{white}
& \cellcolor{green!15}\textbf{0-shot} 
& \cellcolor{green!15}\textbf{MT$_\text{Train}$} 
& \cellcolor{green!15}\textbf{MT$_\text{Train}$+GPT-4o}  
& \cellcolor{green!15}\textbf{All\_LLM} 
& \cellcolor{green!15}\textbf{LLM\_Filt} 
& \cellcolor{yellow!20}\textbf{0-shot} 
& \cellcolor{yellow!20}\textbf{MT$_\text{Train}$} 
& \cellcolor{yellow!20}\textbf{MT$_\text{Train}$+GPT-4o}  
& \cellcolor{yellow!20}\textbf{All\_LLM} 
& \cellcolor{yellow!20}\textbf{LLM\_Filt} \\
\midrule
Qwen 2.5 & 70.5 / 15.6 & 70.7 / 18.3 & 71.0 / 18.1 & \textbf{71.6 / 18.8} & 70.6 / 15.4 
         & 69.9 / 14.0 & \textbf{70.9 / 17.8} & \textbf{70.9 / 18.1} & 71.3 / 16.7 & 70.2 / 13.7 \\
Llama 3.1 & 59.7 / 4.4 & \textbf{72.1 / 21.4} & 72.0 / 20.7 & 71.8 / 20.4 & 71.3 / 18.8 
          & 59.3 / 4.2 & \textbf{71.4 / 19.2} & 71.2 / 18.6 & 71.3 / 17.7 & 68.8 / 15.7 \\
Gemma 2 & 69.7 / 14.5 & \textbf{72.3 / 22.5} & 72.1 / 21.5 & 71.9 / 20.9 & 71.4 / 19.7 
        & 69.4 / 14.1 & \textbf{72.4 / 21.3} & 72.1 / 20.7 & 71.9 / 19.0 & 70.4 / 17.3 \\
SahabatAI Llama & 59.2 / 9.3 & \textbf{72.1 / 22.2} & 60.2 / 8.9 & 62.5 / 12.4 & 64.6 / 12.8 
               & 58.1 / 7.9 & \textbf{71.5 / 20.5} & 59.9 / 8.0 & 62.7 / 10.9 & 70.4 / 16.6 \\
SEA-LION Llama & 69.7 / 16.9 & \textbf{72.4 / 21.3} & 72.0 / 20.5 & 71.9 / 20.7 & 71.6 / 19.6 
               & 69.7 / 15.4 & \textbf{72.1 / 19.5} & 71.5 / 18.0 & 71.5 / 18.0 & 71.1 / 18.5 \\
\bottomrule
\end{tabular}%
}
\end{table*}

\begin{table}[t]
\centering
\caption{Human evaluation results for Javanese and Sundanese.}
\resizebox{\linewidth}{!}{%
\begin{tabular}{llcccc}
\toprule
\textbf{Language} & \textbf{Model} & \textbf{Coherence} & \textbf{Fluency} & \textbf{Correctness} & \textbf{Cultural Rel.} \\
\midrule
\multirow{2}{*}{\cellcolor{green!15}\parbox[c][2.5\baselineskip][c]{1.4cm}{\centering Javanese}} 
    & Gemma 2         & 4.92 & 4.94 & 76 & \textbf{92} \\
    & SEA-LION Llama  & \textbf{4.98} & \textbf{4.98} & \textbf{84} & \textbf{92} \\
\midrule
\multirow{2}{*}{\cellcolor{yellow!20}\parbox[c][2.5\baselineskip][c]{1.4cm}{\centering Sundanese}} 
    & Gemma 2         & 4.22 & 4.74 & 74 & 86 \\
    & SEA-LION Llama  & \textbf{4.40} & \textbf{4.92} & \textbf{78} & \textbf{94} \\
\bottomrule
\end{tabular}}
\label{tab:human_eval}
\end{table}

\subsubsection{Human Evaluation} We conducted a human evaluation to compare models between SEA-LION Llama and Gemma 2 fine-tuned on All LLM-generated data, the two models that showed strong performance in both classification and generation tasks (see Table~\ref{tab:human_written_accuracy},~\ref{tab:mt_test_accuracy},~\ref{tab:human_written_bertscore_rl}, and ~\ref{tab:mt_test_bertscore_rl}). Using the human-written test set, annotators were presented with a story premise and the generated ending sentence. They were asked to rate the outputs based on coherence, fluency, correctness, and cultural relevance, following the same guidelines as in the earlier manual evaluation. As shown in Table~\ref{tab:human_eval}, both models perform well in Javanese and Sundanese, with SEA-LION Llama slightly outperforming Gemma 2 across most human evaluation criteria. However, this advantage is less evident when measured using automatic metrics.

\section{Conclusion}
We explored the potential and limitations of LLM-generated data for commonsense reasoning and story generation in Javanese and Sundanese, introducing the first cloze dataset for these languages with high-quality test sets. Our preliminary analysis in classification and generation settings shows that GPT-4o and Claude-3 Opus demonstrate strong capabilities in generating plausible short stories but face challenges in fluency and cultural accuracy. Despite these limitations, our findings suggest that LLM-assisted data generation is a practical and effective approach for constructing datasets in low-resource languages.


\section{Limitations}
This study acknowledges several limitations in terms of cultural nuance, language scope, and dialectal representation. While we carefully curated data across 12 cultural topics, our focus was limited to only two local languages—Javanese and Sundanese. Although these are the most widely spoken regional languages in Indonesia, they do not capture the full linguistic and cultural diversity present in the world. Moreover, our predefined topics and data sources may not comprehensively reflect the rich variation of cultural practices, dialects, and regional expressions within these languages.



\section{Ethical Considerations}
All human-written datasets have been manually
validated to ensure that harmful or offensive questions are not present in the dataset. We paid our expert workers fairly, based on the monthly minimum wage in Indonesia\footnote{The average monthly minimum wage in Indonesia is approximately 3,000,000 IDR. The workload to complete all the tasks equates to roughly 8 days of full-time work. Each worker was paid 1,250,000 IDR accordingly}. All workers were informed that their stories submitted would be used and distributed for research. Furthermore, no sensitive or personal information about the workers would be disclosed.


\clearpage
\bibliography{anthology,custom}

\clearpage 
\appendix
\label{section:appendix}

\section{Training Configurations}
\label{sec:training_configurations}

For classification, we set the maximum token length for the pre-trained language model to 450 for the premise and 50 for the ending sentence. The model was trained over 20 epochs with early stopping (patience set to 5), using a batch size of 40, Adam optimizer, an initial learning rate of 5e-6 for XLM-R, and a warm-up phase comprising 10\% of the total training steps.

For training the text generation models, we use 4-bit quantization with a LoRA rank of 64 and a LoRA alpha of 128. The models are trained with a batch size of 8, gradient accumulation of 8, a learning rate of 2e-4, and for a single epoch. We employ the \texttt{Unsloth.ai} framework for efficient fine-tuning \citep{unsloth}.

For training the classifier for data filtering, we fine-tune the XLM-R model with a maximum token length of 1024 for the premise and 128 for the ending sentence. The model was trained over 26 epochs, using a batch size of 16, Adam optimizer, an initial learning rate of 1e-5 for XLM-R, and a warm-up phase comprising 5\% of the total training steps.

\begin{figure}[ht]
    \centering
    \includegraphics[width=\linewidth]{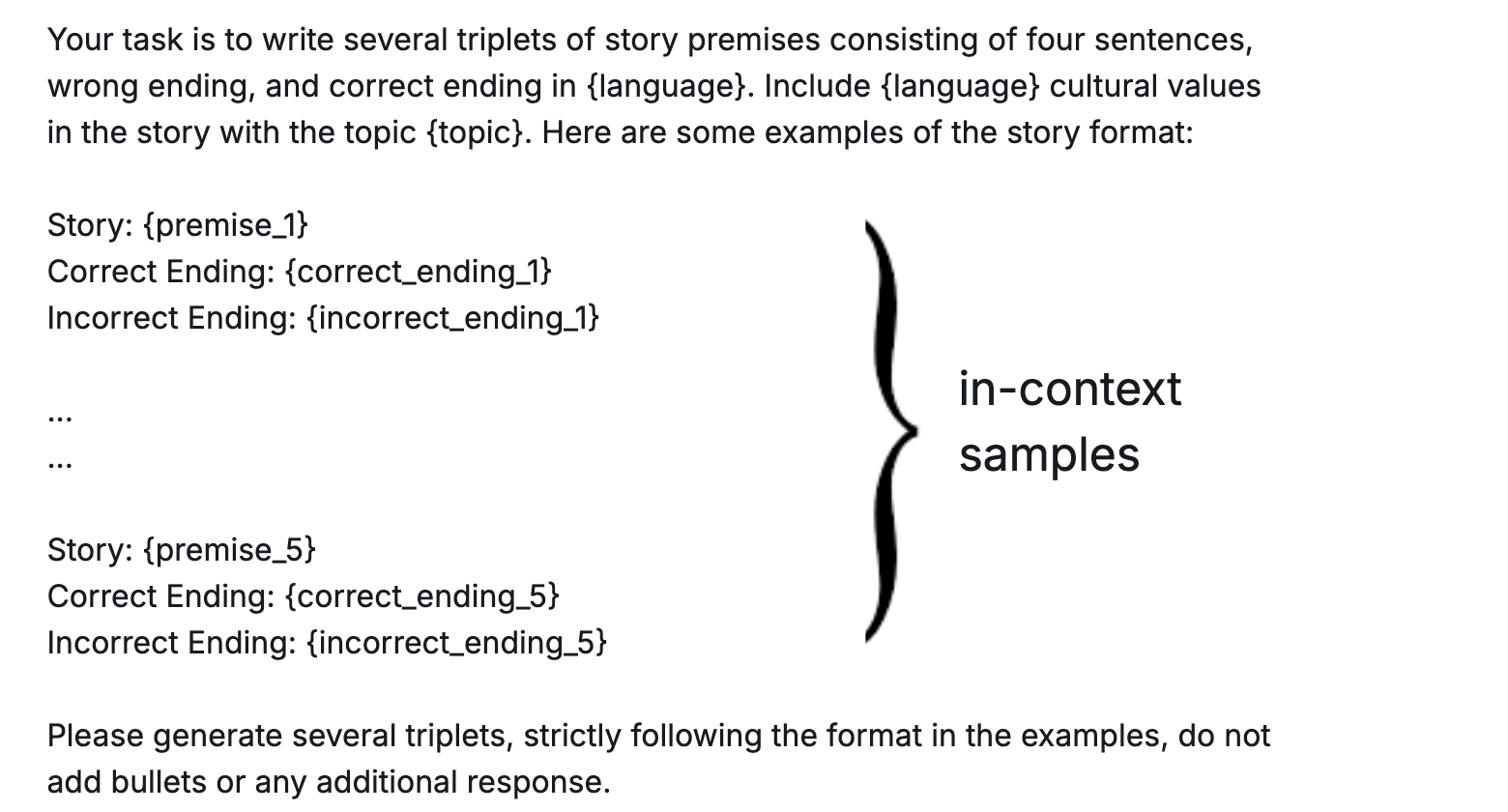}
    \caption{Prompt template instructing LLM to generate a new example for Javanese and Sundanese.}
    \label{fig:prompt_template}
    \vspace{-0.5cm}
\end{figure}

Figure~\ref{fig:prompt_template} shows the prompt template used for in-context learning, guiding the LLM to generate new Javanese and Sundanese.

\section{Distribution of Filtered LLM-Generated Training Data}
\label{sec:filtered_data_proportion}
Initially, training data was generated using six different LLMs, with each model contributing approximately 16.67\% of the total 10K samples (around 2K samples per model). However, after filtering the bad examples, the final dataset composition shifted. The data consist of 1,220 samples, the distribution is as follows: Claude (37.7\%), GPT-4o (29.0\%), LLama (8.6\%), Cohere (4.3\%), and Gemma-2 (1.7\%).

\label{sec:prompt_template}

\section{Agreement Score}
\label{sec:agreement_score}
We measure the fluency and coherency agreement scores using Pearson's correlation and computed the correctness and cultural relevance scores using Cohen's kappa. They are summarized in Table~\ref{tab:agreement_scores}.

\begin{table}[ht]
\centering
\small
\caption{Agreement Scores for Javanese and Sundanese}
\resizebox{0.7\linewidth}{!}{
\begin{tabular}{lcc}
\hline
\textbf{Metric} & \textbf{Javanese} & \textbf{Sundanese} \\ \hline
\textbf{Fluency}           & 0.73 & 0.76 \\ 
\textbf{Coherency}         & 0.77 & 0.71 \\ 
\textbf{Correctness}       & 0.74 & 0.50 \\ 
\textbf{Cultural Relevance} & 0.75 & 0.49 \\ \hline
\end{tabular}}
\label{tab:agreement_scores}
\vspace{-0.2cm}
\end{table}



\section{Workers Scoring Guidelines}
\label{sec:worker_guideline}
\begin{itemize}
    \item \textbf{Fluency (0--5):} Each sentence should be grammatically correct and fluent.
    \begin{itemize}
        \item \textbf{5:} All sentences are grammatically correct and fluent.
        \item \textbf{0:} Sentences are grammatically incorrect and lack fluency.
    \end{itemize}
    
    \item \textbf{Coherency (0--5):} The story should be coherent, with all sentences logically connected. 
    \begin{itemize}
        \item \textbf{5:} Story is highly coherent, with clear and logical flow between sentences.
        \item \textbf{0:} Sentences are disconnected and lack a logical sequence.
    \end{itemize}
    
    \item \textbf{Correctness (Binary):} The correct story closure should be valid, while the incorrect closure should clearly be wrong.
    \begin{itemize}
        \item \textbf{1:} The correct ending is indeed correct, and the incorrect ending is clearly wrong.
        \item \textbf{0:} Either the correct ending is not valid, or the incorrect ending is not clearly wrong.
    \end{itemize}
    
    \item \textbf{Cultural Relevance (Binary):} The story should reflect appropriate cultural norms, values, or symbols relevant to the language.
    \begin{itemize}
        \item \textbf{1:} The story contains relevant cultural norms, values, symbols for the corresponding language.
        \item \textbf{0:} The story lacks cultural relevance or includes irrelevant cultural aspects.
    \end{itemize}
\end{itemize}

\textbf{Scoring Guidelines:}
\begin{itemize}
    \item \textbf{Coherency:} Ranges from 0 to 5, where 5 means each sentence is strongly connected and flows well with the previous and next sentence.
    \item \textbf{Fluency:} Ranges from 0 to 5, where 5 indicates all sentences are grammatically sound and highly fluent.
    \item \textbf{Correctness:} A binary score of 0 or 1 to ensure that the correct ending is truly valid and the incorrect ending is clearly wrong.
    \item \textbf{Cultural Relevance:} A binary score of 0 or 1 to ensure the whole story contains appropriate and relevant cultural symbols and norms for the language being used.
\end{itemize}

\section{Topic and Story-Writing Guidelines}
\label{sec:story_guideline}
We create a total of 300 native-authored stories as part of the Javanese and Sundanese Cloze Project. These stories are evenly distributed across 12 predefined topic categories, with 25 stories per topic. Each story must reflect traditional Javanese and Sundanese values and customs, with attention to detail and coherence in the narrative. The topic categories and their subcategories follows IndoCulture \cite{koto-etal-2024-indoculture}. 
Each story must consist of 4 sentences and two endings: one correct and one incorrect. Ensure that all stories adhere to the topics and categories outlined in the taxonomy and reflect traditional values and cultural relevance.

\section{Machine Translation Evaluation}
\label{sec:machine_translation_quality_comparison}
\begin{table}[h]
\centering
\small
\begin{tabularx}{\columnwidth}{lcc}
\hline
 & \textbf{Javanese} & \textbf{Sundanese} \\
\hline
Google Translate & 0.11 / 0.41 / 45.2 & 0.09 / 0.35 / 42.8 \\
NLLB-200-3.3B    & 0.18 / 0.30 / 53.1 & 0.12 / 0.33 / 43.8 \\
\hline
\end{tabularx}
\caption{BLEU F1 / METEOR / ChrF comparison of Google Translate and NLLB Dense Transformers Translation Model \cite{costa2022no} for Indonesian to Javanese / Sundanese translation on NusaX dataset \cite{winata-etal-2023-nusax}}
\label{tab:translation_metrics}
\end{table}

\end{document}